\documentclass[11pt,a4paper]{article}
\usepackage[hyperref]{acl2020}
\usepackage{multirow}
\usepackage{times}
\usepackage{latexsym}
\usepackage{breqn}

\usepackage{graphicx}
\usepackage{appendix}
\usepackage{graphicx}
\usepackage{subcaption}
\DeclareUnicodeCharacter{2212}{-}

\usepackage{microtype}

\aclfinalcopy % Uncomment this line for the final submission

\begin{document}
\title{KEYword based Sampling (KEYS) for Large Language Models}

\author{
  Jyothir S V$^1$, Zuhaib Akhtar$^1$ \\
  New York University\\
  \texttt{\{sja491,}
  \texttt{za2023}
  \texttt{\}}
  \texttt{@nyu.edu}
}

% \date{}
\maketitle

\begin{abstract}
Question answering (Q/A) can be formulated as a generative task \cite{mitra2017generative} where the task is to generate an answer given the question and the passage (knowledge, if available). 
Recent advances in QA task is focused a lot on language model advancements and less on other areas such as sampling\cite{krishna2021hurdles}, \cite{nakano2021webgpt}. Keywords play very important role for humans in language generation. (Humans formulate keywords and use grammar to connect those keywords and work). In the research community, very little focus is on how humans generate answers to a question and how this behavior can be incorporated in a language model. In this paper, we want to explore these two areas combined, i.e., how sampling can be to used generate answers which are close to human-like behavior and factually correct. Hence, the type of decoding algorithm we think should be used for Q/A tasks should also depend on the keywords. These keywords can be obtained from the question, passage or internet results. We use knowledge distillation techniques to extract keywords and sample using these extracted keywords on top of vanilla decoding algorithms when formulating the answer to generate a human-like answer. In this paper, we show that our decoding method outperforms most commonly used decoding methods for Q/A task.

\end{abstract}

\section{Introduction}

Generating \footnote{Authors contributed equally}text from Language Models requires us to decode (sample) words. To sample words at each time step, various text decoding algorithms are used such as greedy, beam search, temperature based, top-k, nucleus, etc. They have some downsides especially when the task is Question Answering. This is because when humans form answers to a question, they first gather keywords they know about that question and the subject using the knowledge they have and then formulate the answer. For example, If someone asks “What does Professor Bowman do?”, we’ll think of keywords such as "Professor", "New York University", "Natural Language Processing", "GLUE", etc and then use these keywords to finally formulate the answer. Mostly everything else in the answer such as grammar are just fillers.\\
\cite{harabagiu-etal-2000-experiments} did information retrieval and generation using keywords for Question Answering task and this motivated us to explore keyword based decoding for human-like answer generation.\\
Our work shows that this decoding is done to simulate how humans generate answers to a question, which gives better scores for Q/A task. We focus on keywords, as language model gives distribution over vocabulary and to directly sample from that distribution is not an ideal approach when we are dealing factual Q/A. For the question, "What is the capital of Australia?". Language models would generate distribution over cities such as "Sydney", "Perth", "Canberra", etc. But the factually correct keyword that should be sampled is "Canberra". Keyword based sampling (KEYS) would re-weight the probability of "Canberra" as it would be present in the context for that question which makes our answer factually correct.\\
This decoding method also introduces trustworthiness in the generated answer as it restricts the model to generate tokens that are allocated in it's knowledge and domain and doesn't include tokens which are not. If we carefully choose our knowledge base, then it will force the model to pick keywords from that knowledge base rather than generating keywords which are harmful, incorrect or biased. For example,for question "What is the capital of Australia?", knowledge base will have words such as  ”Sydney”, ”Perth”, ”Canberra”, etc. But count of ”Canberra” in the in the article about "capital of Australia" will have large number of keywords of ”Canberra” than ”Sydney”, ”Perth”. It ensures that ”Canberra” has higher weight than ”Sydney”, ”Perth” when the distribution is re-weighted. To ensure numerical stability, we normalized the count scores for each keywords by the highest frequency word.

\section{Related Work}
\begin{figure*}
  \centering
  \includegraphics[scale=0.4]{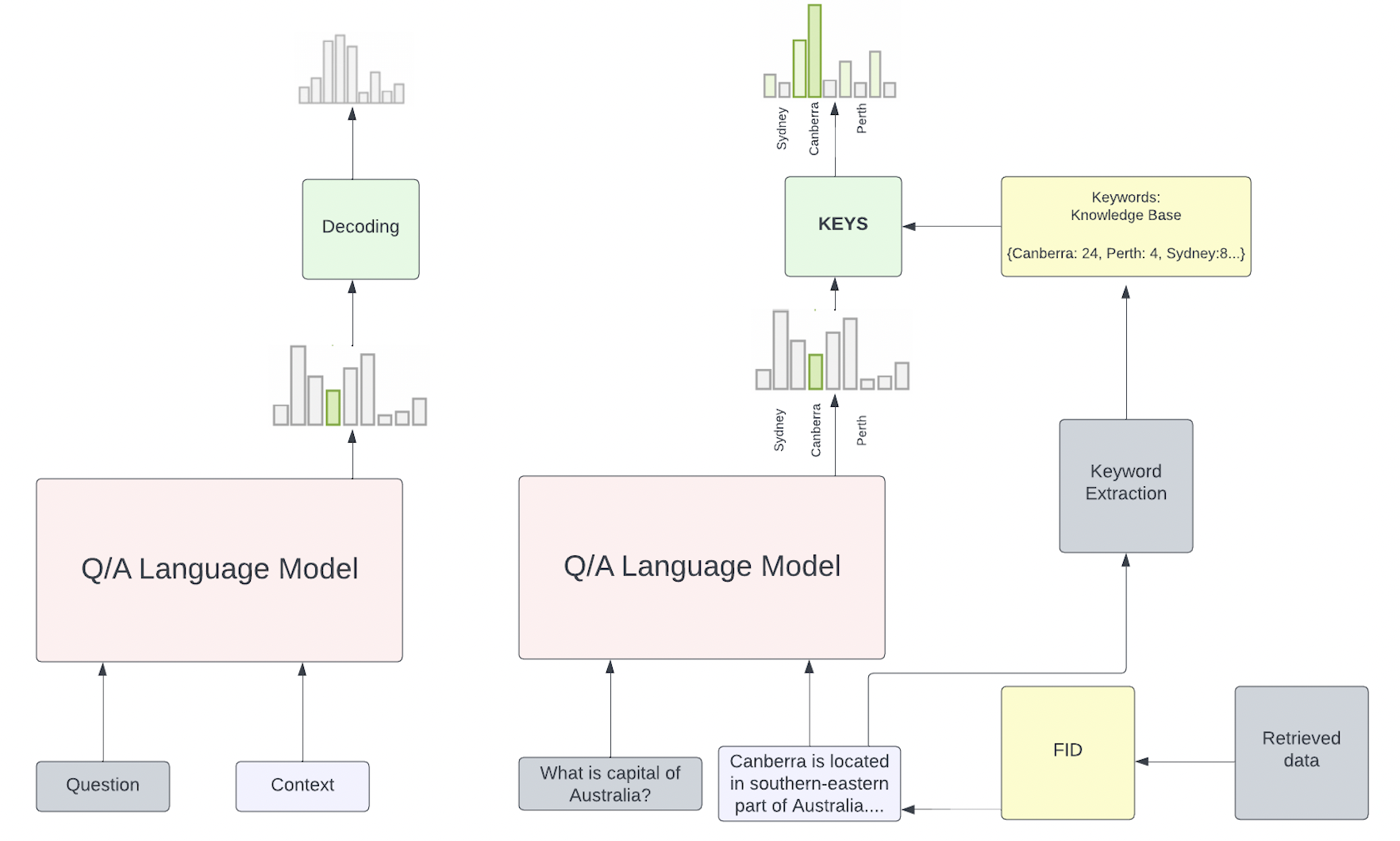}
  \caption{A comparison between a general decoding method (on the left) and Keyword based decoding (KEYS, on the right). KEYS re-distributes the probability over the tokens according to the knowledge base about the question. This generates answer which are factual as it uses keywords and text which is reliability, robustness, and safe even though the Q/A Language model may be biased and untrustworthy.}
\end{figure*}

There are number of sampling methods that are used in generative task such as greedy, beam search, temperature, top-k, nucleus sampling. Greedy sampling suffers from repetitions when decoding is done. Beam search is good but computationally expensive. Temperature based sampling \cite{hinton2015distilling} introduces a hyper-parameter in the distribution where higher values mean softer distribution. So the hyper parameter is meant to increase the probability for probable words for being picked and reduces for less probable words. The problem is that it doesn’t focus on special “keywords” we want to focus on while decoding. Nucleus sampling \cite{holtzman2019curious} takes the idea of top-k sampling one step further such that the smallest possible ‘k’ words are chosen such that the summation of their probability is greater than some threshold. Hence, it focuses on very certain words model thinks is correct but it still fails to work for our hypothesis where focus should be given to special “keywords” too. Hence, we propose a sampling method which will help to prove/disprove the hypothesis.\\

As shown by \cite{durmus2020feqa}, the generated text from a language model suffers from unfaithfulness  which means that the generated text is inconsistent from the source text such as passage. These inconsistencies are not captured in existing metrics. Faithfulness is calculated by generating summaries of source text (human annotations). Language models suffer trade-off between attractiveness and faithfulness. In ideal case, we need both of them to be high. But they suffer due to systemic errors like random sampling \& hallucinations. This is main idea in the introduction of Faithfulness Metric FEQA (Faithfulness Evaluation with Question Answering). This new metric gives us human annotated faithfulness scores. 

\cite{shuster2022language} shows language models generate factually incorrect (but fluent) answers. Methods such as document aggregation to counter this problem were proposed but they result in factually incorrect response. SeeKeR Language Model uses a modular approach where one of the modules is to search the internet to generate the knowledge which is then used to generate the text. This approach outperforms state-of-the-art model according to human ratings of consistency which is an important method to generate answers which are more human-like. Hence, we will use one of the module to generate knowledge and sample keywords from this to generate outputs according to human ratings of consistency and thus will test our method to be more human-like.

\cite{indurthi2017generating} shows a fascinating problem of generating question answer pairs from a given knowledge graph could have be used for several downstream tasks. To extract keywords from entities and relationships and finally generate a natural language question that has a unique answer; such language models which generate questions from keywords performed better (state-of-the-art) than  template based methods for generating questions which motivates us to generate answer in this way which is in the form of keywords where attention is given to those keywords which is also is more human-like behavior.

\begin{table*}[ht]\small
\caption{Evaluation of various metrics for BART Langugae model on different decoding strategies. Results show that our KEYS sampling performs better than various decoding strategies. }
\centering
\begin{tabular}{{|c|c|c|c|c|c|c|c|}}
\hline
Decoding Method& rogue 1& rogue 2 & rogue L & RogueLSum & Bleu & BertScore & BartScore\\
\hline
Temp 0.5   & \textbf{0.249}    &0.0203&  0.141 & \textbf{0.182}&  0.0147 & 0.861 & -3.16 \\
Temp 0.75  & 0.149    &0.016&   0.115 & 0.139&  0.0062 & 0.844 & -3.34 \\
TopP 0.9   & 0.224   &0.018&   0.134 & 0.162&  0.0086 & 0.868 & -3.20 \\
TopP 0.95  & 0.199    &0.015&   0.118 & 0.142 &  0.0071 & 0.852& -3.30 \\
TopK 30    & 0.221    &0.0191&  0.134 & 0.155&  0.0093 & 0.847& -3.19 \\
TopK 40    & 0.158    &0.012&   0.095 & 0.112&  0.0043 & 0.815& -3.41 \\
KEYS + Temp       & 0.244  & \textbf{0.0216}&   \textbf{0.145} & 0.177 & \textbf{0.0161} & \textbf{0.878} & \textbf{-3.02} \\
\hline
\end{tabular}
\end{table*}

\section{Methodology}

\subsection{Dataset}
Most common form of Questioning Answering task is answer sentence selection or reading comprehension where a distribution over the passage is generated to answer a question \cite{rajpurkar2018know}. But to study how human generates answers, we should shift the focus to rather generate an answer to a question. Such type of task is actually relevant for our study. ELI5 (Like I’m Five) Question Answering Dataset is also similar to MS Marco but it is long-form question answering where it has a task which requires to elaborate and in-depth answers to openended questions. This dataset consists of 270K threads from Reddit where people provide answers to questions which are understood easily. It contains Q/A pairs that are up to July 2018 and only those questions were taken which has more upvotes than downvotes.Also, 100 most relevant web searches were captured for each question excluding reddit. Again, this dataset perfectly aligns with our study to find how human generates answers to a question.

\subsection{KEYS Decoding}
    
 We tried most common decoding algorithms such as greedy, beam search, temperature \cite{hinton2015distilling}, top-k, nucleus sampling \cite{holtzman2019curious}, etc and our keyword based sampling layer on top of most of the vanilla decoding algorithms and compare the results.

We experimented with two hyper-parameters of KEYS. One hyper-paramter "lambda" re-weights the distribution of the language model according to the keywords (decide the decoding weight on extracted keywords). Higher the value, higher will be the influence of keywords during decoding from the language model.\\
The other is Keyword-history overlap. It checks the various length overlaps between keywords (of various lengths) and the text generated texts from the langauge models. It then re-weights the distribution based on the degree of overlap and assigns higher weight to the subsequent portion of that group of keyword. Higher the value, higher weight will be assigned and higher the chances for it to appear of the output from the language model.

\begin{align}
P(x | x_{1:i-1}) =
 \begin{cases}
  P(x|x_{1:i−1}) * \alpha_x & if x \in K(w), \notag\\
  P(x | x_{1:i-1}), & otherwise \notag
 \end{cases}
\end{align}

\textit{where $w$ is keyword and $K(w)$ is a lookup table with corresponding words and it's count in the knowledge base and $x$ belongs to $K(w)$}\\

\begin{align}
K(w) = [W1:C1, W2:C2,...,W_k:C_k]\notag\\
\alpha_x \in [C1/C_h,C2/C_h,...,C_K/C_h]\notag
\end{align}
We trained BART language model using maximum likelihood. The size of total question answer pair is $n$. Then,
%\begin{equation}
\begin{align*}
	P(X_1, X_2, \cdots, X_n) = P(X_1)\cdot P(X_2|X_1)\cdot \\
 P(X_3|X_1, X_2) \cdots P(X_n|X_{n-1}, X_{n-2}, \cdots, X_1) \\
 = \prod_{i=1}^{n} \prod_{k=1}^{t_i} P(X_{k}|X_{1}^{k-1},Q_i)\notag
\end{align*}	
%\end{equation}

In the above equation $K(s)$ is keywords extracted from context that is relevant for query. where $w$ are keywords, $C_i$ is respective count, $C_h$ is the highest count keyword K(s) is keywords extracted from context that is revelant for query. $C_h$ is a normalizing constant which re-weights extarcted keywords. As we use the count of keywords in our knowledge base to assign the probability to keywords in the distribution. This count ensures that irrelevant keywords in the knowledge base doesn't affect the probability of correct keywords that should be sampled.

\subsection{Keyword extraction and Language Model}
   We use knowledge distillation techniques like fusion in decoder(FiD)\cite{Izacard2020} which uses cross-attention to retrieve information from given data. We collect data from both dataset and internet (bing search API) to get top k relevant text \cite{shuster2022language}.
    
    Keyword extraction from the knowledge is done using rapid automatic keyword extraction (RAKE) which is an unsupervised and language independent method for keyword extraction \cite{rose2010automatic}. As Q/A task is domain independent, and humans generates answers across all domains using their knowledge, hence domain independent keyword extraction method (RAKE) is used. It focuses on words which carry meaning within a document (content words).  As this is a generative task, we’ll feed Question and other text segment (such as passage) and the language model will output answer to that question. Various decoding strategy is applied alongwith KEYS and results are compared in Table 1.
\subsection{Metrices and Tools}
To study such a behavior, Question answering (Q/A) should be formulated as a generative task for our research \cite{mitra2017generative}. This work directly relates to our work as we also propose this problem as answer (text) generation task. Although this paper uses RNN styled seq2seq architecture which are common for summarization and translations task, we’ll be using large language models (LLMs) which are much better in generating text and overcomes the problem faced by RNN styled seq2seq models. This paper also suggests to use ROGUE for this task because it measures how much readability and correctness is preserved in the answers which is very important when answering a question and not just one that gets good lexical similarity metrics. \\
    As we have posed this problem as human like natural language generation, we used ROUGE \cite{lin-2004-rouge}, Bleu \cite{papineni-etal-2002-bleu}, BertScore \cite{,zhang2019bertscore} and BartScore  \cite{yuan2021bartscore} to evaulate it. This is because these criteria measure correlation with human generated answers. Our objective here is to measure human-like behavior when generating answers.\\
 Hugging face has variety of language models and datasets which we have used in our experiments. They can be directly accessed by their API. 
We used ELI5 Question Answering Dataset \cite{fan2019eli5}  as they have human generated answers (all publicly available). Hence, it will help prove our hypothesis on how humans formulate answers.

% \section{Model Architecture} 
% [TBD]

% \section{Evaluation Setup}

\section{Results and Future Work}
The results from our experiments are shown above in Table 1. It shows that our method outperforms almost all other decoding/sampling methods when generating human-like answers to a question i.e., it performs better than 5 of the 7 decoding strategies. Most recent metrics such as BertScore and BartScore proves that KEYS decoding is better at generating human-like answers. As generating answers is similar to generating responses from chatbots, we will be applying KEYS to chatbots. It makes sense as chatbots should act like personal assistant which needs human-like behaviour and metrics which evaluates them should be proxies to human judgements \cite{khayrallah2023choose}. \\

\section{Acknowledgements}
We want to thank to Prof. Samuel Bowman for his invaluable guidance during the Natural Language Understanding course, where the initial conceptualization of this work was undertaken as a class project.

\bibliographystyle{acl_natbib}
\bibliography{proposal_citations}

\end{document}